\documentclass[letterpaper, 10pt, conference]{ieeeconf}
\IEEEoverridecommandlockouts                       
\usepackage{microtype}
\usepackage{graphicx}
\usepackage[font=small]{caption} 
\usepackage[font=small]{subcaption}
\usepackage{booktabs}
\usepackage[hidelinks]{hyperref}
\usepackage{listings}
\usepackage{float}
\usepackage{amsmath,amsfonts,amssymb}

\newtheorem{problem}{Problem}

\usepackage{yaacro}
\usepackage{tabularx}
\usepackage{cite}
\usepackage{soul}
\usepackage{algpseudocode}
\usepackage[colorinlistoftodos]{todonotes}

\usepackage[utf8]{inputenc}


\usepackage[ruled, vlined, linesnumbered]{algorithm2e}

\floatname{algorithm}{Procedure}

\newcommand\footnoteref[1]{\protected@xdef\@thefnmark{\ref{#1}}\@footnotemark}

\renewcommand{\baselinestretch}{0.974}

\newcommand{\fixlabels}[1]{\textcolor{purple}{[FIX THE LABELS!]}}

\newcommand{\argmaxnew}{\mathop{\mathrm{arg max}}\limits}

\title{\LARGE \bf Communication-Constrained Multi-Robot Exploration\\with Intermittent Rendezvous
}

\author{Alysson Ribeiro da Silva$^{1}$, Luiz Chaimowicz$^{1}$, Thales C. Silva$^{2}$, and M. Ani Hsieh$^{2}$%
\thanks{
This work was supported by ARL DCIST CRA W911NF-17-2-0181 and Office of Naval Research (ONR) Award No. N00014-19-1-2253.
This work was also partially supported by CAPES, CNPq, and Fapemig.
}
\thanks{Alysson Ribeiro da Silva and Luiz Chaimowicz are with the VeRLab, Universidade Federal de Minas Gerais, Brazil. Thales C. Silva and Ani Hsieh are with the GRASP Laboratory, University of Pennsylvania, USA. 
        {\tt\small \{alysson.silva,chaimo\}@dcc.ufmg.br}, {\tt\small \{scthales,mya\}@seas.upenn.edu}.
         } }

\begin{document}
\maketitle
\thispagestyle{empty}
\pagestyle{empty}

\begin{abstract}

Communication constraints can significantly impact robots' ability to share information, coordinate their movements, and synchronize their actions, thus limiting coordination in Multi-Robot Exploration (MRE) applications. 
In this work, we address these challenges by modeling the MRE application as a DEC-POMDP and designing a joint policy that follows a rendezvous plan. 
This policy allows robots to explore unknown environments while intermittently sharing maps opportunistically or at rendezvous locations without being constrained by joint path optimizations.
To generate the rendezvous plan, robots represent the MRE task as an instance of the Job Shop Scheduling Problem (JSSP) and minimize JSSP metrics.
They aim to reduce waiting times and increase connectivity, which correlates to the DEC-POMDP rewards and time to complete the task.
Our simulation results suggest that our method is more efficient than using relays or maintaining intermittent communication with a base station, being a suitable approach for Multi-Robot Exploration.
We developed a proof-of-concept using the Robot Operating System (ROS) that is available at: {\small\url{https://github.com/multirobotplayground/ROS-Noetic-Multi-robot-Sandbox}}.

\end{abstract}


\section{Introduction}



The exploration of unknown environments by a team of robots is an important task for applications like search and rescue, environmental monitoring and patrolling, and mapping of unknown environments.
Robots must define goals, navigate, and collect and share information in a distributed manner \cite{Bramblett2022} to accomplish this efficiently.
However, multi-robot exploration (MRE) requires coordination among team members and is often conducted in geometrically complex environments where robots are subject to constraints on their sensing, mobility, and communication capabilities.


There is a large body of work in the multi-robot coordination domain where the objective is to maintain a connected communication network at all times as the team executes various tasks in both known \cite{Jensen2018, Ong2021, Nonsmooth2023} and unknown environments \cite{Pratissoli2022,Zhang2022,ZhangAndHao2022}.
However, keeping the robots connected can restrict exploration, especially when team sizes and communication ranges are small, leading to increased exploration times.
Differently, intermittent communication strategies for teams of robots have been employed in various application settings \cite{Hollinger2012,Kantaros2019,Hannes2020}.
These works suggest that MRE strategies that employ intermittent communication can be more performative since they allow the team to cover more space, possibly in less time \cite{Bramblett2022}.
To allow intermittent rendezvous or encounters, the authors in \cite{Hollinger2012} and \cite{Hannes2020} propose formulating a navigation plan through joint-path planning using connectivity graphs \cite{Bullo2009}.
Similarly, Kantaros et al.~\cite{Kantaros2019} propose an intermittent communication method that divides robots into sub-teams, and members in different sub-teams must rendezvous periodically following a predetermined schedule.
In their method, robots update their paths given their next encounter to satisfy a global \textit{Linear Temporal Logic} (LTL) formula every time they meet.
Unfortunately, these approaches are infeasible for more general MRE applications in which the environment is unknown and paths must adapt as robots explore.

\begin{figure}[t!]
    \centering
    \includegraphics[width=0.4\textwidth]{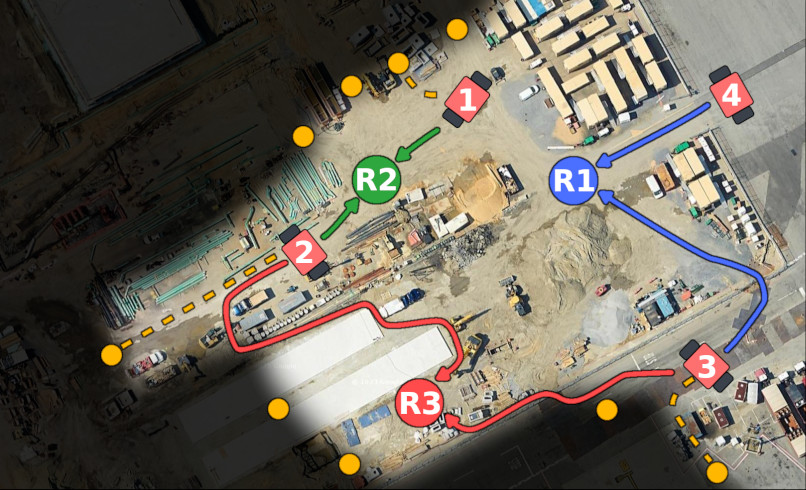}
    \caption{Robots meeting at rendezvous locations spread over a construction site. The R1, R2, and R3 are hypothetical rendezvous locations. Colored arrows indicate the rendezvous locations. Orange circles are places to explore and dashed lines represent the paths followed by robots.}
    \label{fig:schematics}
\end{figure}

In this work, we propose a different approach where we model the MRE task as a Decentralized Partially Observable Markov Decision Process (DEC-POMDP) \cite{Matignon2012}.
The objective is to design a policy considering intermittent communication that maximizes its reward related to exploring new places and sharing maps. 
Due to the large policy search space of the MRE task, we propose an induced joint policy composed of robot controllers.
Each controller allows a robot to select promising places to explore and, at the same time, attend to agreed rendezvous locations scheduled in a rendezvous plan ({\it e.g.}, Fig.~\ref{fig:schematics}).
To ensure that robots meet the time constraints imposed by data collection and exploration sub-tasks, we propose representing the rendezvous plan as an instance of the Job Shop Scheduling Problem (JSSP) \cite{Gao2019,Fang2019, Song2023}.
A significant advantage of employing instances of the JSSP is the ability for robots to coordinate tasks and estimate the resources needed to complete them while minimizing the time they wait at rendezvous locations due to navigation and exploration uncertainties.

Our contributions can be summarized as follows:

\begin{itemize}
    \item We represent the MRE task as an instance of the JSSP to enable the generation of a rendezvous plan for intermittent communication. The mapping of the MRE problem into a JSSP enables the minimization of wait times at rendezvous locations, maximizing the returns of the DEC-POMDP, and results in rendezvous encounters that can be updated during exploration.
    \item We propose a decentralized joint policy for the MRE application  based on a DEC-POMDP, where each robot's controller aims to maximize the utility of visiting frontiers while considering the rendezvous plan to share maps. Unlike existing methods, our approach enables robots to explore freely and attend to rendezvous locations only when necessary, eliminating the necessity for joint-path optimization among team members.
\end{itemize}

We evaluated our method considering that robots have a short communication range and considered scenarios where the workspace's length scale is large to avoid incidental all-time connectivity when exploring.
Due to the vast MRE literature, we design two baselines, based on \cite{Bramblett2022} and \cite{Jensen2013}, using a distributed frontier-based method that delivers information to a base station and maintains a relay network.
We measured the time taken to complete the exploration and the rewards of our method's DEC-POMDP against the baselines in our simulator in different urban scenarios obtained from satellite images.
Our results suggest that our method is able to explore faster and obtain higher rewards for the underlying DEC-POMDP without a base station or a relay network.
In addition, we provide a proof of concept deployment using Gazebo Simulator and the Robot Operating System (ROS).
\section{Background and Problem Statement}
\label{sec:problem_statement}



We define a set of $n$ robots as $\mathcal{R}$, where each one explores an unknown bounded environment $\Omega$. A robot $i$ perceives the environment, builds a local map, estimates its pose, and constructs the frontier set at each time step $t$, where frontiers are the boundary between known and unknown regions. It then explores by maintaining its current state $s_i(t)$, represented by a tuple with the relative poses of nearby robots, its local map, and its pose. An action $a_i(t) \in \mathcal{A}_i(t)$ consists in navigating towards a location in $\Omega$. For the MRE application, an action location refers to a frontier or a rendezvous location ({\it i.e.}, point of encounter between two or more robots). We define the utility of an action as $u_i(a_i(t))$, which reflects the benefit of navigating towards the location specified by $a_i(t)$ considering its path cost. To coordinate exploration among multiple robots, a task allocation algorithm \cite{Bramblett2022} helps selecting frontiers that maximize the sum of utilities.

We model the connectivity among robots at a time step $t$ as a graph $\mathcal{G}(t) = (\mathcal{R}, \mathcal{E}(t))$, where an edge $e(t) = (i, j)$ from $\mathcal{E}(t)$ represents the connectivity between robots $i$ and $j$. We use a communication model based on a communication range. Therefore, at each time step $t$, an edge $e(t)$ between robots $i$ and $j$ exists only if they are close to each other. Importantly, we also define a sub-team as a set of two or more robots $\mathcal{T}_s$ that periodically meet.
%
%
Robots are intermittently connected if the union of the graphs $\{\mathcal{G}(t_a),\ldots,\mathcal{G}(t_b)\}$ forms a connected graph for every $t_b-t_a$ steps. When robots are connected, they share and merge their maps to enhance efficiency. Robots engage in exploration by asynchronously selecting actions and executing them until there are no more frontiers left on their local map


The underlying DEC-POMDP of the MRE application is defined as the tuple $(\mathcal{R}, \mathbb{S}, \mathbb{A}, T, \mathbb{O}, O, R, h)$, where $\mathbb{S}$ is a set of all possible joint states, $\mathbb{A}$ is a set of joint actions, $T$ is the set of conditional transition probabilities, $\mathbb{O}$ is the set of joint observations, $O$ is the set of observations probabilities, $R$ is the immediate reward, and $h$ is the horizon of the problem ({\it i.e.}, time to finish the exploration). For simplicity, we omitted the initial system's conditions. We define a joint state $s(t) \in \mathbb{S}$ as $(s_1(t),\ldots,s_n(t))$ and a joint action $a(t) \in \mathbb{A}$ as $(a_1(t),\ldots,a_n(t))$, where $\mathbb{A} = \mathcal{A}_1 \times \ldots \times \mathcal{A}_n$. A joint observation $o(t) \in \mathbb{O}$ is defined as $o(t) = s(t)$ at a time step $t$, and its probability is $P(o(t)|a(t),s(t+1))$. The transition and observation probabilities are usually unknown for the MRE application, and robots can not measure the true state of the system due to localization and mapping uncertainties.

The immediate reward $R : \mathbb{A} \times \mathbb{S} \rightarrow \mathbb{R}$ is function that associates actions and states to real numbers, and we define the \textit{return} of the DEC-POMDP as a bounded sum $\sum^{h}_{t=0} R(s(t), a(t))$ because $\Omega$ has a limited number of frontiers. For the MRE application, we approximate the reward $R(s(t), a(t))$ as the sum of the utility of the actions $\sum^{n}_{i=1} u_i(a_i(t))$ from the joint action $a(t)$. In this regard, we define a \textit{joint policy} as $\pi = \{\pi_1,...\pi_n\}$, where $\pi$ is decentralized, and each $\pi_i$ is a finite state controller \cite{Amato2013} that aims at maximizing the return by taking appropriate actions. From the perspective of a decentralized policy, robots are unaware of the joint state $s(t)$, joint action $a(t)$, joint observation $o(t)$, and the return is available only when searching for a policy and before exploration takes place.

\begin{problem}[MRE Problem with Intermittent Rendezvous]
Design a set of decentralized robot controllers $\pi = \{\pi_1,...,\pi_n\}$, where each controller $\pi_i$ directs robot $i$ to explore frontiers through joint actions $a(t)$. Ensure that the union of the connectivity graphs $\{\mathcal{G}(t_a),\ldots,\mathcal{G}(t_b)\}$ is connected for every $t_b-t_a$ steps until there are no more frontiers to explore. The objective is to maximize the return of the DEC-POMDP, defined as $\sum^{h}_{t=0}\sum^{n}_{i=1} u_i(a_i(t))$, where $a_i(t)$ represents the action of robot $i$ in the joint action $a(t)$. At each time step, robots operate based on their states $s_i(t)$ and are unaware of the joint state $s(t)$, joint actions $a(t)$, and joint observations $o(t)$, and the return is available only during the planning phase.
\end{problem}

\begin{figure*}[t!]
    \centering
    \includegraphics[width=0.99\textwidth]{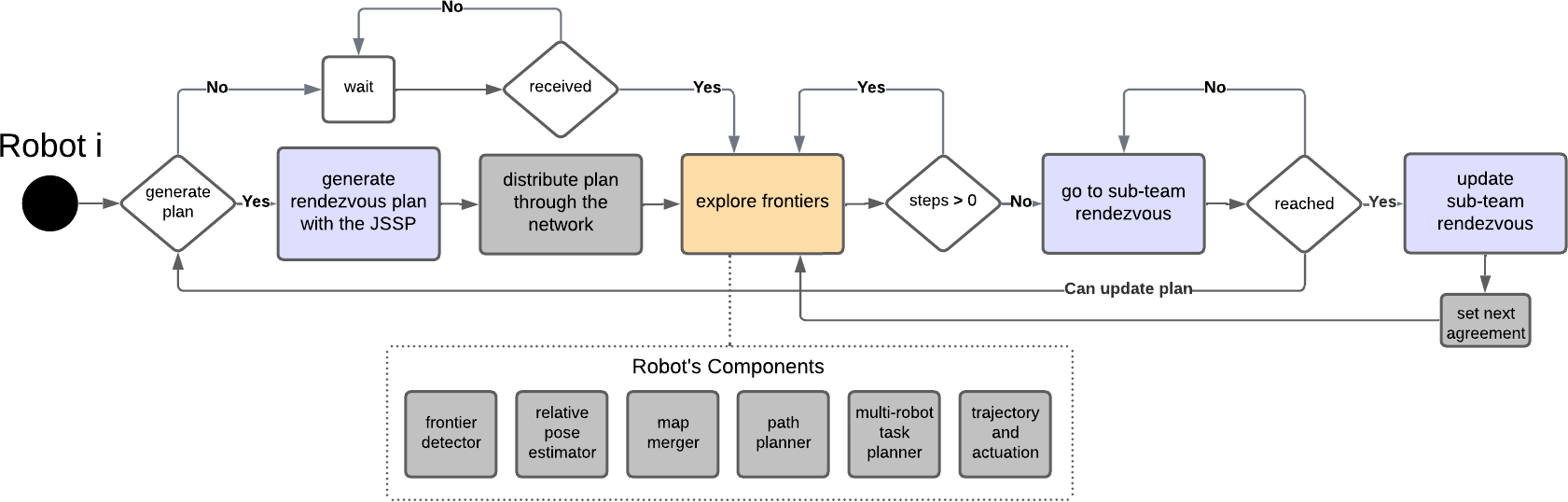}
    \caption{Diagram for robot $i$ controller integrated with the rendezvous plan generator and the robot's components. The purple components generate a rendezvous plan through a JSSP, decide when and with whom robots should meet, and update the rendezvous locations. Differently, the light orange component refers to the exploration task and it makes robots explore unknown places. The robot's components are used for deployment and help assemble the robot's state and execute actions.}
    \label{fig:execution}
    \vspace{-2.0em}
\end{figure*}


\begin{figure}[t!]
    \centering
    \includegraphics[width=0.45\textwidth]{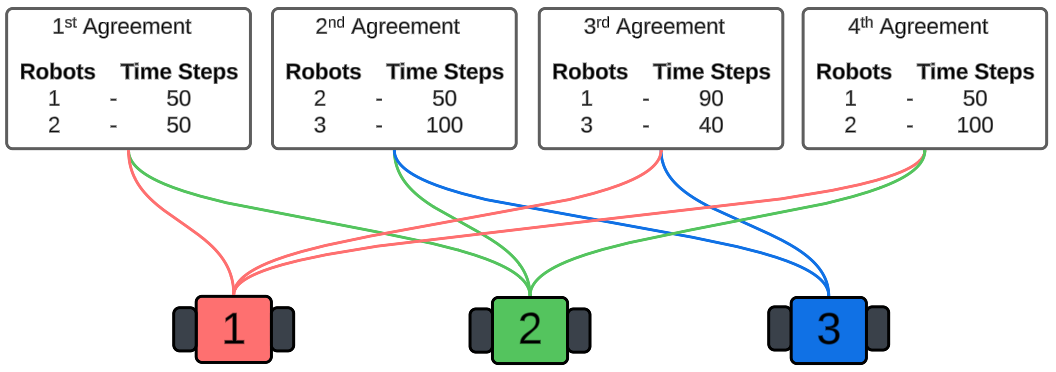}
    \caption{Hypothetical agreements between $3$ robots (red, green, blue). Robot $1$ participates in the $1^{st}$, $3^{rd}$, and $4^{th}$ agreements, while robot $2$ participates in the $1^{st}$, $2^{nd}$, and $4^{th}$ ones. In the $1^{st}$ agreement, robots $1$ and $2$ should explore for $50$ time steps before going to a meeting location to share maps.}
    \label{fig:agreements}
\end{figure}

\section{Methodology}
\label{sec:method}

Our solution to Problem 1 has two main parts: an exploration \textit{plan} that defines sub-team structures and resource allocation for the exploration, and a decentralized joint policy that controls navigation to rendezvous locations and frontiers.
Before exploring, one robot from $\mathcal{R}$ calculates an initial rendezvous plan for the team, considering an estimate about the area of $\Omega$ and the robots' sensing and traveling speed.
Upon receiving their plans, the robots begin exploration by executing their local policy, which selects actions based on the current state.

%

%
%
%

%
%

We define the plan through agreements between robots, where agreements specify which sub-team $\mathcal{T}_s \subseteq \mathcal{R}$ of robots should meet and how much each robot must explore before going to an assigned rendezvous location. We use time steps as a metric of how much resources a robot should spent before meeting with a sub-team.
%
%
To fulfill an agreement, all robots from its sub-team must meet at its rendezvous location and share maps.
Robots update their sub-team rendezvous location and, for simplicity, they can update the plan if the encounter is of a sub-team composed by all robots.

%
%
%

%
%

%
%

%
%

Our solution is illustrated in Fig~\ref{fig:execution}.
In summary, the robots will first create an exploration plan. 
They will then follow their local policy to explore the frontiers based on predefined agreements, which specify the time they should spend exploring and which sub-team they should meet next. 
Finally, the robots will navigate to one of their sub-teams rendezvous locations in a specific order, share maps, define a new rendezvous location for the current sub-team, and resume exploring. 
We explain the specific details of the process in the following sections and provide a pseudo-code in Algorithm~\ref{alg:smainloop}. 
In particular, in Section \ref{subsec:rep_rendezvous_plan} and \ref{subsec:generating_kw}, we present our method to generate the plans, and in Section \ref{sec:explore}, we present the policy for exploration.

\subsection{Representing the Rendezvous Plan}
\label{subsec:rep_rendezvous_plan}

We propose an Agreement matrix $K$ and a Steps matrix $W$ to represent the plan efficiently.
Each element of $K$ is a binary number, and each element of $W$ is a positive real. 
The rows of both matrices holds information about the agreements, and columns represent robots' index. 
For example, the following $K$ and $W$ instances represent the agreements from Fig.~\ref{fig:agreements},


\vspace{-1.0em}
\begin{align*}
K = \begin{bmatrix}
        1 & 1 & 0\\
        0 & 1 & 1 \\
        1 & 0 & 1 \\
        1 & 1 & 0
    \end{bmatrix}
\label{eq:matrices} &&
W = \begin{bmatrix}
        50 & 50  & 0\\
        0  & 50  & 100 \\
        90 & 0   & 40 \\
        50 & 100 & 0
    \end{bmatrix}
\end{align*}

\noindent
where, the first {row of $K$ tells that the first rendezvous of robot $1$ should be with robot $2$, whereas the second row tells that the first rendezvous of robot $3$ should be with robot $2$. 
Since agreements are fulfilled in order, this example implies that robot $3$ is free to explore while robots $1$ and $2$ fulfill their first agreement.
The $W$ matrix tells how many time steps each robot should explore before fulfilling an agreement from $K$. 
For instance, the first row of $W$ tells that robots $1$ and $2$ must explore for $50$ time steps before meeting.
The structure of matrices $K$ and $W$ also inform us the number of sub-teams in $\mathcal{R}$.
%
In this example there are three sub-teams, where $\mathcal{T}_1=\{1,2\}$, $\mathcal{T}_2=\{2,3\}$, and $\mathcal{T}_3=\{1,3\}$} the rendezvous locations are associated with the sub-teams, that can meet at different moments and locations.

\subsection{Generating K and W}
\label{subsec:generating_kw}

While $K$ and $W$ serve as compact representations of agreements, it is crucial to define plans that effectively coordinate the robots' activities, balancing exploration and travel to rendezvous locations.
This coordination minimizes waiting times and prevents resource waste. 
Consequently, to generate the matrices, we propose transforming $K$ and $W$ into a JSSP instance $\mathcal{I}$, changing the instance parameters by optimizing metrics such as the \textit{Makespan} ({\it i.e.,} the total time to finish a set of sub-tasks) and obtaining the optimized $K$ and $W$ from $\mathcal{I}$.
%
%
%
%
A JSSP instance $\mathcal{I}$ represents the execution of the plan through time, and it is composed of machines (robots), where each machine has a list of jobs to execute.
It is important to note that jobs are sub-tasks of exploring the environment associated with starting and ending times.
In this regard, robots explore the environment when executing jobs and fulfill one agreement for each job they finish.

In this work, we particularly designed a \textit{Genetic Algorithm} \cite{Zou2024} solver with a population of $(K,W)$ pairs, initialized randomly, to generate the plan, because it is not trivial to find gradients to iterate the plan over the fitness landscape. The solver evaluates the population by converting each pair into a JSSP instance $\mathcal{I}$ and extracting metrics we present in Section~\ref{sec:metrics}.
Algorithm~\ref{alg:jssp} shows the transformation of each individual $(K,W)$ into $\mathcal{I}$, which creates jobs by iterating over rows of $K$.
At the same time, it uses the values from the steps matrix $W$ and the length of the jobs already inserted into $\mathcal{I}$ to adjust their starting and ending times.
The algorithm marks jobs from the same agreement at line $8$ to help recover $K$ and $W$ after the optimization.

\subsection{JSSP Instance Metrics}
\label{sec:metrics}

We designed the following objective function to evaluate a JSSP instance $\mathcal{I}$ obtained from a $(K,W)$ pair,

\begin{equation}
z(\mathcal{I}) = \sum_a \alpha_ah_a(\mathcal{I}) + \sum_b \beta_bg_b(\mathcal{I})
\end{equation}

\noindent
where, $h_a$ and $g_b$ are metrics obtained from $\mathcal{I}$, $\alpha_a$ and $\beta_b$ are weights. 
The intuition behind $z(\mathcal{I})$ is to minimize the Makespan and waiting times during rendezvous encounters and it is represented by the following metrics:

\begin{itemize}
    \item \textbf{Plan Error:} $\sum\alpha_ah_a(\mathcal{I})$ is used to generate plans considering an estimate of how big is $\Omega$ with minimum waiting times, a fulfillment frequency, and the Makespan.
    \item \textbf{Connectivity Metrics:} $\sum\beta_bg_b(\mathcal{I})$ is used to check if the robots are going to be intermittently connected if they follow the plan.
\end{itemize}

\begin{algorithm}[t!]
\footnotesize
\DontPrintSemicolon
\KwIn{Number of robots $|\mathcal{R}|$\\
        \quad\quad\quad Rendezvous agreement matrix $K$\\
        \quad\quad\quad Rendezvous exploration steps matrix $W$}
\KwOut{JSSP instance $I$}
\SetKwBlock{Begin}{function}{end function}
\SetKwRepeat{Do}{do}{while}
\Begin($\text{JSSP} {(|\mathcal{R}|, K, W)}$) {
    Let $\mathcal{I}$ be a JSSP instance with $|R|$ machines;\;
    \ForEach{$agreement$ row in $K$} {
        \ForEach{$robot$ column in $K[agreement]$} {
            \If{$K[agreement][robot] = 1$} {
                create a new job $job_{new}$ and allocate it to the $jth$ machine in $\mathcal{I}$;\;
                set the length of $job_{new}$ as $W[agreement][robot]$ and update its starting and ending times according to what was already allocated to the $jth$ machine;\;
                mark $job_{new}$ as belonging to $agreement$ to help recovering $K$ and $W$ through the inverse process;\;
            }
        }
    }
    \Return $\mathcal{I}$;\;
}
\caption{\small{Job-shop Scheduling Transformation}}
\label{alg:jssp}
\end{algorithm}

\subsubsection{Plan Error}
\label{sec:planerr}

From the JSSP instance $\mathcal{I}$, the algorithm calculates the exploration estimate error,

\begin{equation}
\label{eq:workdone}
h_1(\mathcal{I}) = ((\Omega_A - \Omega_E) - \sum_{\text{job}_j \in J_{\mathcal{I}}}job_j\mathcal{X})^2
\end{equation}

\noindent
where, $(\Omega_A - \Omega_E)$ is an estimate of how much area is left to explore, given its estimated size $\Omega_A$, and the area robots already explored $\Omega_E$, $job_j$ is the length of the $jth$ job, and $\mathcal{X}$ is a constant that estimates how much area robots can explore, given the sensor they use to acquire information from $\Omega$ ({\it e.g.}, lidar, sonar, camera) and traveling speed; the work done error,

\begin{equation}
\label{eq:desired}
h_2(\mathcal{I}) = (\sum_{\text{job}_j \in J_{\mathcal{I}}}job_j - \sum_{\text{job}_j \in J_{\mathcal{I}}}(job_j + \mathcal{Y}(job_j)))^2
\end{equation}

\noindent
where, $job_j$ is the job length and $\mathcal{Y}(job_j)$ is the number of time steps that should be added to reduce waiting times at rendezvous locations; and the deviation between rendezvous encounters,

\begin{equation}
\label{eq:sigma_agreement}
h_3(\mathcal{I}) = \sqrt{\frac{1}{|\mathcal{J}_{\mathcal{I}}|}\sum_{job_j \in \mathcal{J}} (job_j - \bar{job})^2}
\end{equation}

\noindent
where, $job_j$ is the job length, $|\mathcal{J}_{\mathcal{I}}|$ is the number of jobs from $\mathcal{I}$, and $\bar{job}$ is the average job length. Importantly, we estimate the size of the environment $\Omega_A$ as a the area of the bounding rectangle that encloses the entire mission zone.


\subsubsection{Connectivity Metrics}
\label{sec:connecterr}

Let $\mathcal{H}_{\mathcal{I}}$ be a set of connectivity graphs obtained from a JSSP instance $\mathcal{I}$.
We propose creating a new graph $\mathcal{G}_{dense}$, which is a condensed version of all graphs in $\mathcal{H}_{\mathcal{I}}$ to calculate the connectivity metrics.
Each vertex from $\mathcal{G}_{dense}$ is a robot, and an edge exists between robots $i$ and $j$ from $\mathcal{G}_{dense}$ if any of the connectivity graphs from $\mathcal{H}(\mathcal{I})$ connects them.
From $\mathcal{G}_{dense}$, we propose extracting the connectivity index,

\begin{equation}
\label{eq:connectivity_index}
g_1(\mathcal{I}) = 1 - \frac{1}{1 + \mathcal{P}(\mathcal{G}_{dense})}
\end{equation}

\noindent
where, $\mathcal{P}(\mathcal{G}_{dense})$ is a function that returns the length of the biggest connected component of $\mathcal{G}_{dense}$. We also propose extracting,

\begin{equation}
\label{eq:edges_count}
g_2(\mathcal{I}) = |\mathcal{E}'|
\end{equation}

\noindent
where, $|\mathcal{E}'|$ is the number of edges from $\mathcal{G}_{dense}$. 
The intuition behind the number of edges is to avoid over-connectivity. 

During the planning phase, it is feasible to estimate the DEC-POMDP's return from the JSSP instance.
This estimation relies on various factors derived from the JSSP, including the time allocated for robot exploration, rendezvous scheduling, and the extent of information exchange during map sharing from eq.~\ref{eq:workdone}.
The optimization process in this context correlates to the horizon of the DEC-POMDP. 
Specifically, minimizing the Makespan implies in a more condensed exploration timeline, leading to a shorter planning horizon within the DEC-POMDP framework.

\begin{algorithm}[t!]
\footnotesize
\DontPrintSemicolon
\KwIn{Agreement matrix $K$\\
    \quad\quad\quad Exploration steps matrix $W$}
\SetKwBlock{Begin}{function}{end function}
\Begin($\text{ExplorationIntermittentRendezvous} {(K, W)}$) {
    Let $\mathcal{T}_{s}$ be the sub-team of the first agreement to fulfill in $K$;\;
    Let $steps$ be the number of steps to explore in $W$ from the agreement of $\mathcal{T}_{s}$;\;
    \ForEach{time step t} {

        Receive relative poses of other robots and their maps if they are within communication range, update local map and merge with the ones received, and estimate my pose;\;
        
        \Comment{$s_i(t)$ is defined as a tuple with the relative poses, the current map, and the robot's pose.}\;
        
        Detect the frontiers set $\mathcal{F}_i(t)$ from $s_i(t)$;\;
        Assemble the connectivity graph $\mathcal{G}_i(t)$ from $s_i(t)$;\;
        
        \If{The frontier set $\mathcal{F}_i(t)$ is empty} {
            \textbf{break};\;
        }

        
        \If(\tcp*[h]{Exploration Actions}){$steps > 0$} {
            \If{Not selected an action} {
                Select an exploration action $a_i(t)$ with a greedy action selection mechanism, considering the single and multi-robot cases;\;
            } \Else {
                Navigate towards $a_i(t)$ location;\;
            }
            $steps = steps - 1$
        } \Else(\tcp*[h]{Agreement Fulfillment Action}) {
            Navigate to the sub-team $\mathcal{T}_{s}$ rendezvous location;\;
            \If{I am in charge of the update \textbf{and} \\
                \quad $\mathcal{G}_i(t)$ connects the sub-team $\mathcal{T}_{s}$} {
                    Calculate the centroids set $\mathcal{C}(t)$ with the KNN from the frontiers in $\mathcal{F}_i(t)$;\;
                    \If(\tcp*[h]{Location update}){$\mathcal{T}_{s} \neq \mathcal{R}$} {
                        Pick a new rendezvous location with a random sampling method from $\mathcal{C}(t)$ and send it to all robots in the sub-team $\mathcal{T}_{s}$;\;
                    } \Else(\tcp*[h]{Synchronization agreement}) {
                         Generate the new rendezvous matrices $K$ and $W$;\;
                         Pick centroids with maximum distance from each other from $\mathcal{C}(t)$, one for each sub-team in the new $K$;\;
                         Send the new matrices and the sub-team rendezvous locations to all robots in $\mathcal{T}_{s}$;\;
                    }
                }
            
                \If{received sub-team rendezvous locations \textbf{or}
                    \quad received plan \textbf{or} timeout} {
                    Set $T_{s}$ to the sub-team specified by the next agreement this robot participates in $K$;\;
                    Update $steps$ with the nex value from $W$ given the next agreement this robot participates in $K$;\;
                }
        }
    }
    Go back to the initial location;\;
}
\caption{\small{Exploration with Intermittent Rendezvous}}
\label{alg:smainloop}
\end{algorithm}


\subsection{Exploration Policy with Intermittent Rendezvous}
\label{sec:explore}

After receiving a rendezvous plan (matrices $K$ and $W$), each robot $i$ explores the environment with its policy $\pi_i$.
%
Robots capture information from the environment $\Omega$ with their sensors and receive observations $o_i(t)$ that they use within their current state.
Based on their state, they decide which action to take. 
An action can be exploring the environment or fulfilling an agreement from $K$.
Importantly, when navigating, robots constantly try to share maps and their relative poses with others within communication range.
%
%

Algorithm~\ref{alg:smainloop} shows how robots gather information, assemble their state, and select their actions.
At each time step, a robot $i$ gathers information from the environment to build and maintain a local map and estimates its pose.
This process involves integrating the observations into the local map and correcting its pose based on sensory input.
It also receives maps, that are merged with its local map, and the relative poses from robots within its communication range.
Finally, a robot represents its state $s_i(t)$ as a tuple containing the relative poses of nearby robots, its local map, and its pose.
Through its state, it calculates the set of frontiers $\mathcal{F}_i(t)$ with a flooding-fill algorithm and the connectivity graph $\mathcal{G}_i(t)$.
Both the frontiers and the connectivity graph are used to navigate and verify the requirements to fulfill agreements.

The policy keeps track of the current state and which agreement to fulfill to select an appropriate action.
The action space $\mathcal{A}_i(t)$ at a time step $t$ of a robot is composed of two types of actions: I) \textit{Exploration Actions}. II) \textit{Agreement Fulfillment Action}.
\begin{itemize}
    \item \textbf{Exploration Actions}: An Exploration Action refers to exploring a frontier, calculating and updating a path towards it given its merged map, and following this path. Robots perform Exploration Actions until they explore for their assigned time to fulfill their current agreement. 
    \item \textbf{Agreement Fulfillment Action}:  Differently, an Agreement Fulfillment Action refers to navigating towards the current rendezvous location, waiting for the assigned sub-team or until a timeout, sharing maps, and selecting the next agreement to fulfill. Robots execute the Agreement Fulfillment Action if they have explored the number of time steps in $W$ specified by the current agreement. After fulfilling the meeting, the assigned sub-team updates its rendezvous location and can update the plan if necessary.
\end{itemize}
Robots decide which type of action to take by verifying the variable \textit{steps} at line $11$ of Algorithm~\ref{alg:smainloop}. They also track which sub-team they should meet when performing an Agreement Fulfillment Action with the variable $\mathcal{T}_{s}$. After executing an action, a robot selects the next agreement it participates in by updating both variables.


\subsubsection{Exploration Actions}

For each frontier in $\mathcal{F}_i(t)$, there is an Exploration Action in $\mathcal{A}_i(t)$.
Basically, a robot estimates the utility of an action as the utility of visiting a frontier $f_j$ from the frontiers set $\mathcal{F}_i(t)$.
It selects an Exploration Action by iterating over all frontiers from $\mathcal{F}_i(t)$ seeking for the one that maximizes the following utility function,

\begin{equation}
\label{eq:frontier_utility}
u_i(f_j) = \frac{\mathcal{N}_{f_j}}{\Phi(f_j)}
\end{equation}

\noindent
where, $\mathcal{N}_{f_j}$ is the area to be explored when navigating towards $f_j$, and $\Phi(f_j)$ is its path cost.


%
The single and multi-robot action selection cases work as follows:

\begin{itemize}

\item{\textbf{Single Robot:}} If a robot can not communicate with others, then it is going to select an action $a_i(t)$ associated with $f_j^* = \argmaxnew_{\forall f_j \in \mathcal{F}_i(t)}u_i(f_j)$.

\item{\textbf{Multi-robot:}} When within communication range, robots from a sub-team have the same merged map and can select which frontier to explore.
In this case, one of the robots selects a joint action using a greedy action selection heuristic that aims to locally maximize the sum of the utilities $u_i(a_i(t))$ for each robot from the sub-team.
This greedy selection also maximizes the DEC-POMDP reward for a joint action $a(t)$.

\end{itemize}

\begin{figure}[t!]
    \centering
    \includegraphics[width=0.38\textwidth]{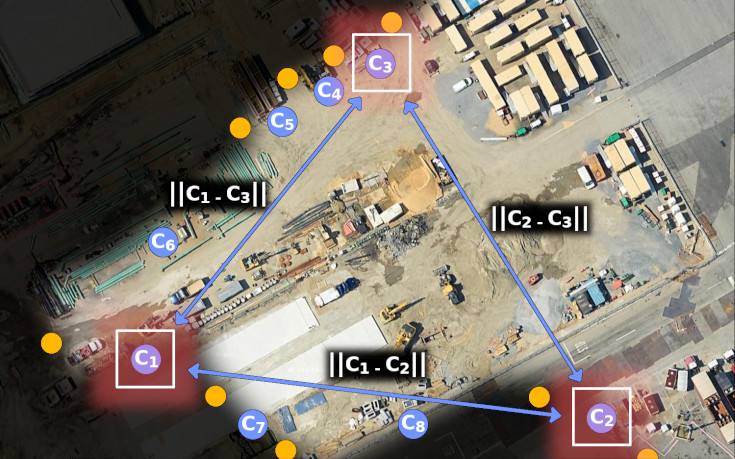}
    \caption{Heuristic to pick new rendezvous locations when robots have common information during synchronization for $3$ sub-teams from the example from Fig.~\ref{fig:agreements} in a construction site. The orange circles are frontiers, and the blue circles are the center of masses obtained from the KNN algorithm. The center of masses $c_1$ and $c_3$ were chosen to the $1^{st}$ and $2^{nd}$ sub-teams. Differently, $c_2$ was chosen to the $3^{rd}$ sub-team by maximizing the distance $\sqrt{(c_1 - c_2)^2 + (c_2 -c_3)^2}$.}
    \label{fig:distances}
\end{figure}

\begin{figure*}[t!]
    \centering
    \includegraphics[width=1.0\textwidth]{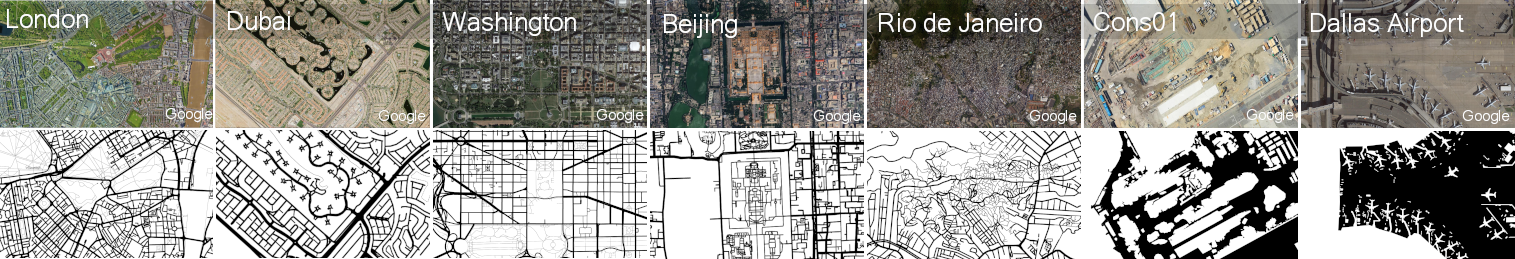}
    \caption{Samples from our benchmark and simulator. (From left to right) Sections of London, Dubai, Washington DC, Beijing, Rio de Janeiro, construction site, and Dallas International Airport.}
    \label{fig:bench_sample}
    \vspace{-1.0em}
\end{figure*}

\begin{table*}[t!]
\centering
\resizebox{1.0\textwidth}{!}{
\begin{tabular}{ clllllllllll }
\multicolumn{1}{l}{} &
  \multicolumn{1}{c}{\centering{NY}} &
  \multicolumn{1}{c}{\centering{DC}} &
  \multicolumn{1}{c}{\centering{Tokyo}} &
  \multicolumn{1}{c}{\centering{Beijing}} & 
  \multicolumn{1}{c}{\centering{Dubai}} &
  \multicolumn{1}{c}{\centering{Moscow}} &
  \multicolumn{1}{c}{\centering{Berlin}} &
  \multicolumn{1}{c}{\centering{London}} &
  \multicolumn{1}{c}{\centering{Paris}} & 
  \multicolumn{1}{c}{\centering{Rio}} \\ \cline{1-11}
\multicolumn{1}{c|}{BS} &
  \multicolumn{1}{l|}{\small${6992\pm 648}$} & 
  \multicolumn{1}{l|}{\small${11849\pm 301}$} & 
  \multicolumn{1}{l|}{\small${10120\pm 719}$} & 
  \multicolumn{1}{l|}{\small${11538\pm 1279}$} & 
  \multicolumn{1}{l|}{\small${14643\pm 1273}$} & 
  \multicolumn{1}{l|}{\small${14613\pm 1117}$} & 
  \multicolumn{1}{l|}{\small${11973\pm 1153}$} & 
  \multicolumn{1}{l|}{\small${11818\pm 1354}$} & 
  \multicolumn{1}{l|}{\small${11399\pm 735}$} & 
  \multicolumn{1}{l}{\small${13871\pm 1213}$} \\ \cline{1-11}  
\multicolumn{1}{c|}{CRN1R} &
  \multicolumn{1}{l|}{\small${6060\pm 377}$} & 
  \multicolumn{1}{l|}{\small${11540\pm 1277}$} &
  \multicolumn{1}{l|}{\small${9579\pm 856}$} &
  \multicolumn{1}{l|}{\small${11209\pm 1126}$} &
  \multicolumn{1}{l|}{\small${13813\pm 1621}$} &
  \multicolumn{1}{l|}{\small${13634\pm 1361}$} & 
  \multicolumn{1}{l|}{\small${10798\pm 1005}$} &
  \multicolumn{1}{l|}{\small${11109\pm 821}$} & 
  \multicolumn{1}{l|}{\small${10826\pm 829}$} &
  \multicolumn{1}{l}{\small${12475\pm 1582}$} \\ \cline{1-11} 
\multicolumn{1}{c|}{CRN2R} &
  \multicolumn{1}{l|}{\small${6398\pm 782}$} & 
  \multicolumn{1}{l|}{\small${10889\pm 549}$} &
  \multicolumn{1}{l|}{\small${9326\pm 614}$} &
  \multicolumn{1}{l|}{\small${10341\pm 980}$} &
  \multicolumn{1}{l|}{\small${13167\pm 1888}$} &
  \multicolumn{1}{l|}{\small${12989\pm 830}$} & 
  \multicolumn{1}{l|}{\small${11729\pm 684}$} &
  \multicolumn{1}{l|}{\small${10558\pm 827}$} &
  \multicolumn{1}{l|}{\small${10359\pm 706}$} &
  \multicolumn{1}{l}{\small${11140\pm 481}$} \\ \cline{1-11} 
\multicolumn{1}{c|}{Ours} &
  \multicolumn{1}{l|}{\small${4607\pm 108}$} &
  \multicolumn{1}{l|}{\small${8048\pm 188}$} &
  \multicolumn{1}{l|}{\small${7056\pm 205}$} &
  \multicolumn{1}{l|}{\small${7174\pm 309}$} &
  \multicolumn{1}{l|}{\small${8061\pm 249}$} &
  \multicolumn{1}{l|}{\small${8297\pm 183}$} &
  \multicolumn{1}{l|}{\small${7632\pm 204}$} &
  \multicolumn{1}{l|}{\small${7760\pm 345}$} &
  \multicolumn{1}{l|}{\small${7436\pm 130}$} &
  \multicolumn{1}{l}{\small${8073\pm 282}$} \\ \cline{1-11} 
\multicolumn{1}{c|}{OursSync} &
  \multicolumn{1}{l|}{\small$\mathbf{4436\pm 117}$} &
  \multicolumn{1}{l|}{\small$\mathbf{7181\pm 218}$} &
  \multicolumn{1}{l|}{\small$\mathbf{6500\pm 399}$} &
  \multicolumn{1}{l|}{\small$\mathbf{6459\pm 274}$} &
  \multicolumn{1}{l|}{\small$\mathbf{7700\pm 413}$} &
  \multicolumn{1}{l|}{\small$\mathbf{7870\pm 316}$} &
  \multicolumn{1}{l|}{\small$\mathbf{6932\pm 117}$} &
  \multicolumn{1}{l|}{\small$\mathbf{6996\pm 116}$} &
  \multicolumn{1}{l|}{\small$\mathbf{7084\pm 170}$} &
  \multicolumn{1}{l}{\small$\mathbf{7460\pm 295}$} \\ \cline{1-11} 
  
\multicolumn{1}{l}{} &
  \multicolumn{1}{c}{\centering{prk01}} &
  \multicolumn{1}{c}{\centering{prk02}} &
  \multicolumn{1}{c}{\centering{prk03}} &
  \multicolumn{1}{c}{\centering{air01}} &
  \multicolumn{1}{c}{\centering{air02}} &
  \multicolumn{1}{c}{\centering{air03}} &
  \multicolumn{1}{c}{\centering{air04}} &
  \multicolumn{1}{c}{\centering{cons01}} &
  \multicolumn{1}{c}{\centering{cons02}} &
  \multicolumn{1}{c}{\centering{cons03}} \\ \cline{1-11}
\multicolumn{1}{c|}{BS} &
  \multicolumn{1}{l|}{\small${10078\pm 881}$} &
  \multicolumn{1}{l|}{\small${2980\pm 285}$} &
  \multicolumn{1}{l|}{\small${7967\pm 744}$} &
  \multicolumn{1}{l|}{\small${5005\pm 656}$} &
  \multicolumn{1}{l|}{\small${10053\pm 1740}$} &
  \multicolumn{1}{l|}{\small${6460\pm 547}$} &
  \multicolumn{1}{l|}{\small${6710\pm 718}$} &
  \multicolumn{1}{l|}{\small${10822\pm 1690}$} &
  \multicolumn{1}{l|}{\small${7003\pm 495}$} &
  \multicolumn{1}{l}{\small${7186\pm 496}$} \\ \cline{1-11}
\multicolumn{1}{c|}{CRN1R} &
  \multicolumn{1}{l|}{\small${9000\pm 1098}$} &
  \multicolumn{1}{l|}{\small${3016\pm 235}$} &
  \multicolumn{1}{l|}{\small${7398\pm 597}$} &
  \multicolumn{1}{l|}{\small${4582\pm 329}$} &
  \multicolumn{1}{l|}{\small${10270\pm 1784}$} &
  \multicolumn{1}{l|}{\small${6306\pm 469}$} &
  \multicolumn{1}{l|}{\small${6120\pm 601}$} &
  \multicolumn{1}{l|}{\small${9262\pm 622}$} &
  \multicolumn{1}{l|}{\small${6379\pm 400}$} &
  \multicolumn{1}{l}{\small${6379\pm 400}$} \\ \cline{1-11}
\multicolumn{1}{c|}{CRN2R} &
  \multicolumn{1}{l|}{\small${9172\pm 802}$} &
  \multicolumn{1}{l|}{\small${2886\pm 165}$} &
  \multicolumn{1}{l|}{\small${6778\pm 512}$} &
  \multicolumn{1}{l|}{\small${4634\pm 497}$} &
  \multicolumn{1}{l|}{\small${9039\pm 868}$} &
  \multicolumn{1}{l|}{\small${6230\pm 661}$} &
  \multicolumn{1}{l|}{\small${6283\pm 682}$} &
  \multicolumn{1}{l|}{\small${8764\pm 644}$} &
  \multicolumn{1}{l|}{\small${6451\pm 302}$} &
  \multicolumn{1}{l}{\small${6388\pm 495}$} \\ \cline{1-11}
\multicolumn{1}{c|}{Ours} &
  \multicolumn{1}{l|}{\small${5833\pm 251}$} &
  \multicolumn{1}{l|}{\small${2242\pm 49}$} &
  \multicolumn{1}{l|}{\small${5054\pm 182}$} &
  \multicolumn{1}{l|}{\small$\mathbf{4310\pm 294}$} &
  \multicolumn{1}{l|}{\small${6985\pm 286}$} &
  \multicolumn{1}{l|}{\small${5197\pm 618}$} &
  \multicolumn{1}{l|}{\small${5427\pm 361}$} &
  \multicolumn{1}{l|}{\small${7034\pm 338}$} &
  \multicolumn{1}{l|}{\small${4954\pm 184}$} &
  \multicolumn{1}{l}{\small$\mathbf{5343\pm 216}$} \\ \cline{1-11}
\multicolumn{1}{c|}{OursSync} &
  \multicolumn{1}{l|}{\small$\mathbf{5458\pm 222}$} &
  \multicolumn{1}{l|}{\small$\mathbf{2226\pm 63}$} &
  \multicolumn{1}{l|}{\small$\mathbf{4895\pm 265}$} &
  \multicolumn{1}{l|}{\small${4436\pm 215}$} &
  \multicolumn{1}{l|}{\small$\mathbf{6538\pm 378}$} &
  \multicolumn{1}{l|}{\small$\mathbf{5023\pm 416}$} &
  \multicolumn{1}{l|}{\small$\mathbf{5263\pm 307}$} &
  \multicolumn{1}{l|}{\small$\mathbf{6553\pm 293}$} &
  \multicolumn{1}{l|}{\small$\mathbf{4717\pm 123}$} &
  \multicolumn{1}{l}{\small${5449\pm 260}$} \\ \cline{1-11}
\end{tabular}
}
\caption{Average number of time steps to finish the exploration for our dataset composed of cities, parking lots (prk), airport sections (air), and constructions sites (cons). }
\label{tab:table_cities}
\vspace{-2.0em}
\end{table*}


\subsubsection{Agreement Fulfillment Action}

A robot tries to fulfill its current agreement after it has explored the assigned number of steps from the plan.
In this regard, it stops executing a previously selected Exploration Action, calculates a path towards the current agreement sub-team $\mathcal{T}_{s}$ rendezvous location, and follows it, as shown at line $18$ of Algorithm~\ref{alg:smainloop}.
After reaching the rendezvous location, robots share their maps and update the location associated with their sub-team.
To select a new rendezvous location, one of the robots from the sub-team creates the locations set $\mathcal{C}(t)$ by clustering frontiers from $\mathcal{F}_i(t)$ with a \textit{K-Nearest Neighbors} (KNN).
Next, it randomly samples a location from $\mathcal{C}(t)$ and sends it to all other robots in the sub-team. 
Finally, the robots from the sub-team select their next agreement and continue with subsequent exploration.

Conversely, to update the rendezvous plan, robots must have common information.
Otherwise, they might generate agreements that can not be fulfilled. 
They achieve this by fulfilling an agreement in which the sub-team has all robots (\textit{i.e.}, $\mathcal{T}_{s} = \mathcal{R}$) which we call a \textit{synchronization agreement}.
%
%
One robot updates the plan through a synchronization agreement at lines $25 - 27$ of Algorithm~\ref{alg:smainloop} after a predetermined number of time steps.
They define the rendezvous location of this agreement as the center of mass of all frontiers they had in common during the last synchronization.
%
%

The robot updating the plan generates new values for $K$ and $W$, which are transmitted to the other robots.
In particular, it exploits the fact that they have common information and assign new rendezvous locations to all sub-teams of the new plan. 
As shown in Fig.~\ref{fig:distances}, it selects the locations from $\mathcal{C}(t)$ with the maximum distance from each other as the new sub-teams rendezvous locations.
This encounter is less harmful for exploration than having all-time connectivity or a base station because communication is majorly intermittent in the plan, which we show through our experiments.
\section{Benchmark Simulation Results}
\label{sec:results}

To evaluate our method, we assembled a dataset composed of $20$ urban environments from satellite images from Google Maps ({\it e.g.}, Fig.~\ref{fig:bench_sample})  and prepared their cell decomposition in advance. 
We executed our approach in our simulator, where robots execute the exploration task in a completely distributed fashion as shown in Fig.~\ref{fig:execution}, plan their paths in the decomposed maps we prepared, and measured the total number of time steps to finish the exploration as done in previous MRE benchmarks [19]. 
In our simulator, robots account only for paths that are executed in discrete time steps.

Our experiments are: I) Benchmark: we benchmark our method against our baselines to evaluate its performance. II) Consistency: we evaluate the spread of the results on $6$ instances of the dataset. III) Larger Mission Areas: we got an instance (London) from our dataset and doubled its dimensions to evaluate the impact of larger areas. IV) Return: we evaluated the return of the DEC-POMDP, as defined in Section~\ref{sec:problem_statement}, in two instances of the dataset (cons01 and London).
We measured the average explored area per time step, and its standard deviation between $10$ runs with $4$ robots for $50000$ steps each, where they have $5$ cells as visibility range and $4$ cells of communication diameter, which resembles low-spec robots. 
At each run, we let the robots start as a cluster in the same location on the map, picked randomly between runs.


We designed three baselines inspired by \cite{Jensen2018} and \cite{Bramblett2022} to compare against our method: a Constrained Static Base Station (BS), a Constrained Relay Network One Relay (CRN1R), and a Constrained Relay Network Two Relays (CRN2R) method.
The base station in the BS and the relays have a communication range of $12$ cells in the cell decomposition.
In the CRN1R, each robot can place $1$ extra relay after $2000$ simulation steps at the communication boundaries with the base station or other relays, while robots in the CRN2R can place $2$ relays.
The comparisons were made against our method (Ours) without updating the plan and our method (OursSync), which performs roughly $4$ synchronizations per mission.
In all baselines, robots must deliver information to the base station or relay network at a fixed frequency.

\subsubsection{\textbf{Benchmark}}

In Table~\ref{tab:table_cities}, we show the average number of time steps robots took to complete the exploration for all $20$ environments in our dataset. 
For city sections, our method took a maximum of $7870$ time steps and a minimum of $4436$ time steps to explore Moscow and New York City, respectively. 
Robots achieved this performance because Moscow is more structured, while New York has many possible routes that are close to each other, facilitating the mission. 
When operating at airports, parking lots, and construction sites, our method took a maximum of $6553$ time steps at cons01 and a minimum of $2226$ time steps at prk02, where cons01 is more complex.
There are no meaningful differences between \textit{Ours} and \textit{OursSync}, which indicates the feasibility of the synchronization without degrading the overall performance.

\subsubsection{\textbf{Consistency}}

\begin{figure}[t!]
    \begin{subfigure}[b]{0.49\columnwidth}
        \includegraphics[width=\textwidth]{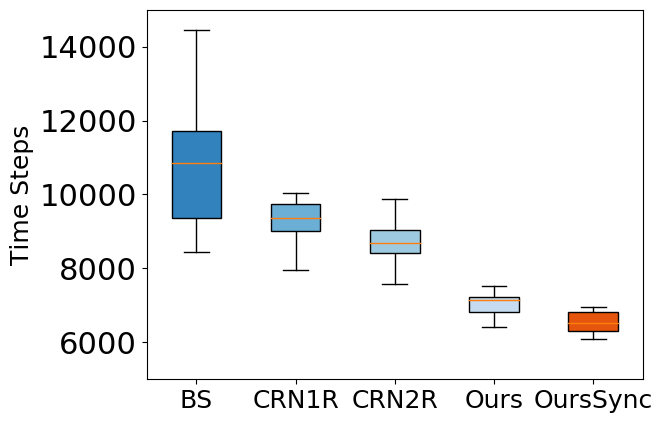}
        \caption{cons01}
    \end{subfigure}
     \begin{subfigure}[b]{0.49\columnwidth}
        \includegraphics[width=\textwidth]{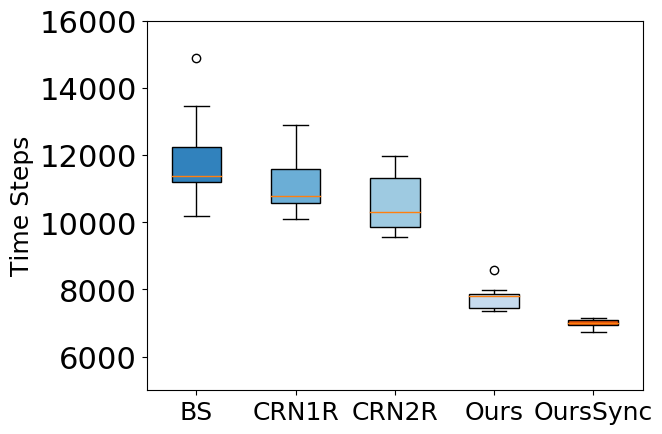}
        \caption{London}
    \end{subfigure}
    \begin{subfigure}[b]{0.49\columnwidth}
        \includegraphics[width=\textwidth]{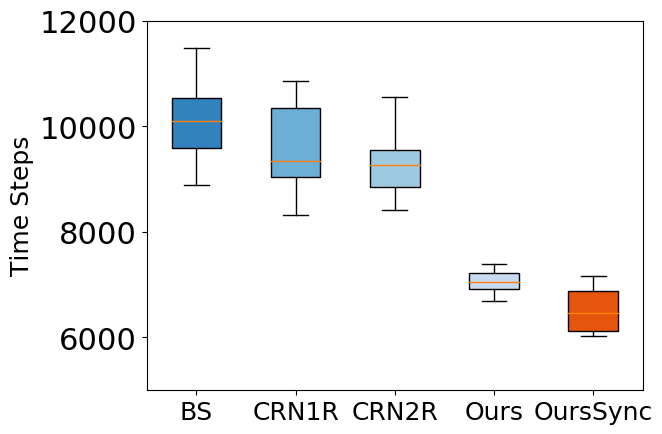}
        \caption{Tokyo}
    \end{subfigure}
     \begin{subfigure}[b]{0.49\columnwidth}
        \includegraphics[width=\textwidth]{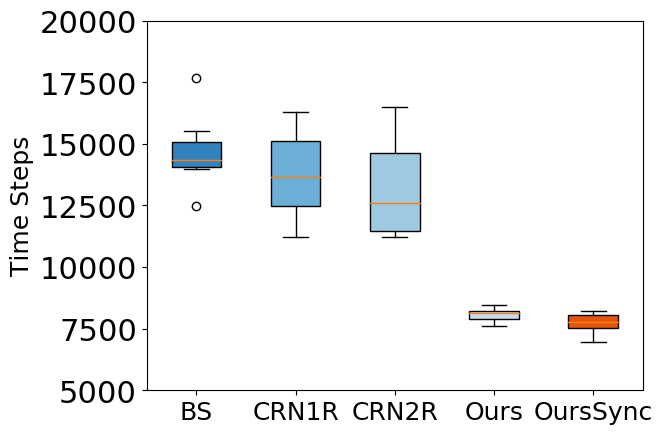}
        \caption{Dubai}
    \end{subfigure}
    \begin{subfigure}[b]{0.49\columnwidth}
        \includegraphics[width=\textwidth]{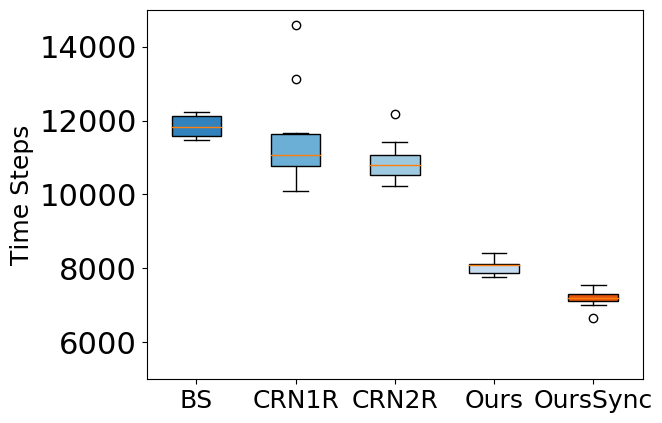}
        \caption{Washington DC}
    \end{subfigure}
     \begin{subfigure}[b]{0.49\columnwidth}
        \includegraphics[width=\textwidth]{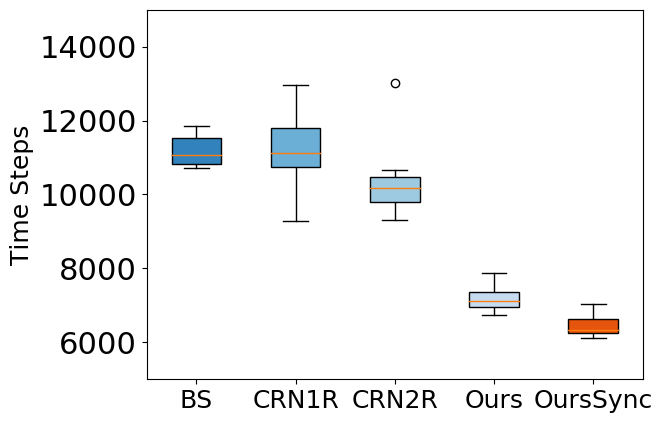}
        \caption{Beijing}
    \end{subfigure}
    \caption{Interquartile range of our baselines against our method. The Y-axis represents simulation steps.}
    \label{fig:interquartile_range}
\end{figure}

We show the spread of the results for different scenarios (cons01, London, Tokyo, Dubai, Washington DC, and Beijing sections) in Fig.~\ref{fig:interquartile_range}. 
The results suggest that our method is more stable with a smaller variation between runs. 
We can enhance the baselines by extending the base station range or adding more relays. 
However, extending this range towards infinity or keeping the robots connected might be impossible, and we expect that our method could achieve even better results for larger areas, as we present next.

\subsubsection{\textbf{Larger Mission Area}}

\begin{figure}[t!]
    \centering

    \begin{subfigure}[b]{0.49\columnwidth}
        \includegraphics[width=\textwidth]{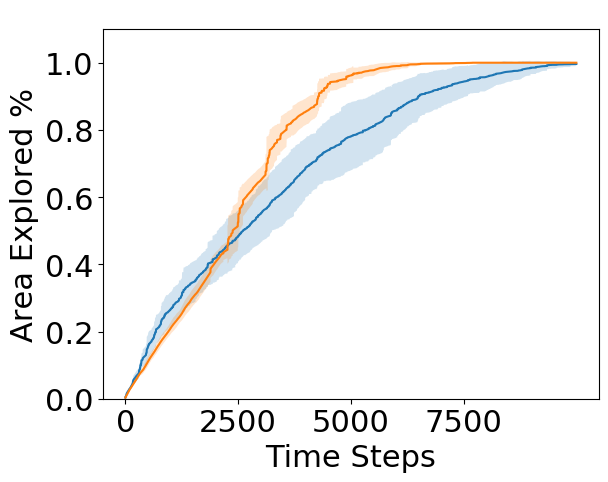}
        \caption{London}
    \end{subfigure}
    \begin{subfigure}[b]{0.49\columnwidth}
        \includegraphics[width=\textwidth]{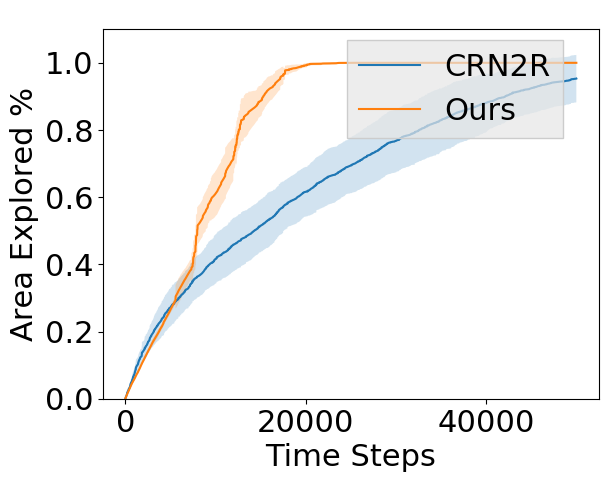}
        \caption{London scaled}
    \end{subfigure}\vspace{0.5em}
    \caption{Percentage of the explored area for each simulation step. The X-axis represents simulation steps, and the Y-axis is the percentage of the area explored.}
    \label{fig:scaled_area}
\end{figure}

Fig.~\ref{fig:scaled_area} compares our method against the CRN2R, which is the best performing baseline, in the London instance from our dataset. 
For this test, we doubled the area of the map in both dimensions. 
These results show that the other methods are constrained by their connectivity, because they have to deliver information to accomplish the mission. 
In contrast, our method obtained gains related to updating the rendezvous location and intermittently sharing maps across the workspace.

\subsubsection{\textbf{Return}}

\begin{figure}[t!]
    \centering
    \begin{subfigure}[b]{0.49\columnwidth}
        \includegraphics[width=\textwidth]{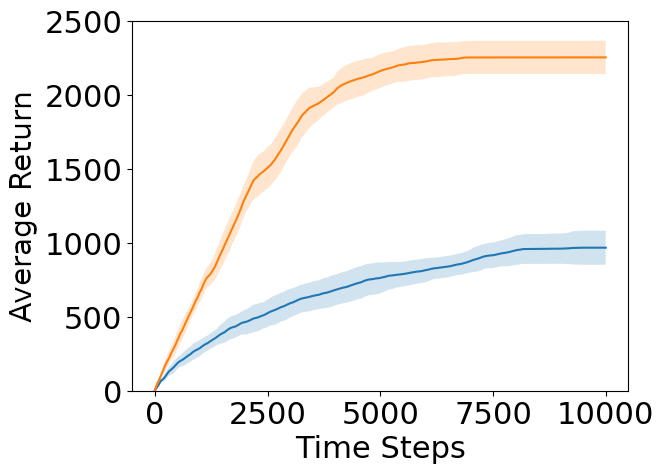}
        \caption{cons01}
    \end{subfigure}
    \begin{subfigure}[b]{0.49\columnwidth}
        \includegraphics[width=\textwidth]{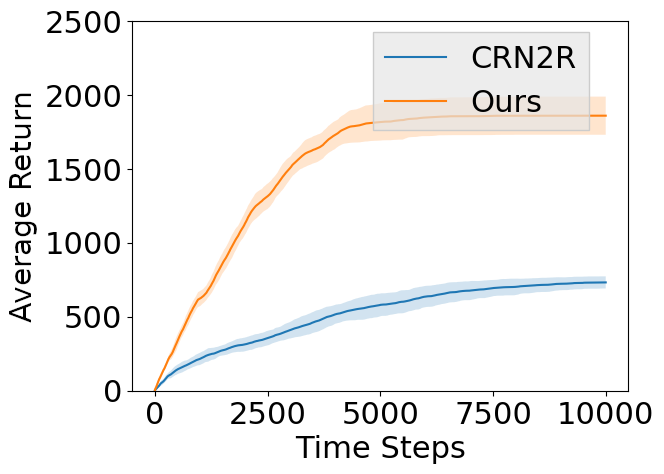}
        \caption{London}
    \end{subfigure}
    \caption{Average return for the DEC-POMDP for each simulation step. The X-axis represents simulation steps, and the Y-axis is the return of the DEC-POMDP as defined in Section~\ref{sec:problem_statement}. We captured the return from the baseline when robots explore frontiers or share maps with others and the relay network.}

    \label{fig:reward_res}
\end{figure}

Fig.~\ref{fig:reward_res} shows the average return of the DEC-POMDP for our methods in cons01 and London sections against the baseline CRN2R.
Our method obtained average returns around $2000$ for both scenarios, while the CRN2R obtained average returns below $1000$.
Our analysis indicates that our method can gather and share more information among robots about the workspace per unit of time while traveling less distances without maintaining a relay network.
%


\begin{figure}[t!]
    \centering
    \begin{subfigure}[b]{0.8\columnwidth}
        \includegraphics[width=\textwidth]{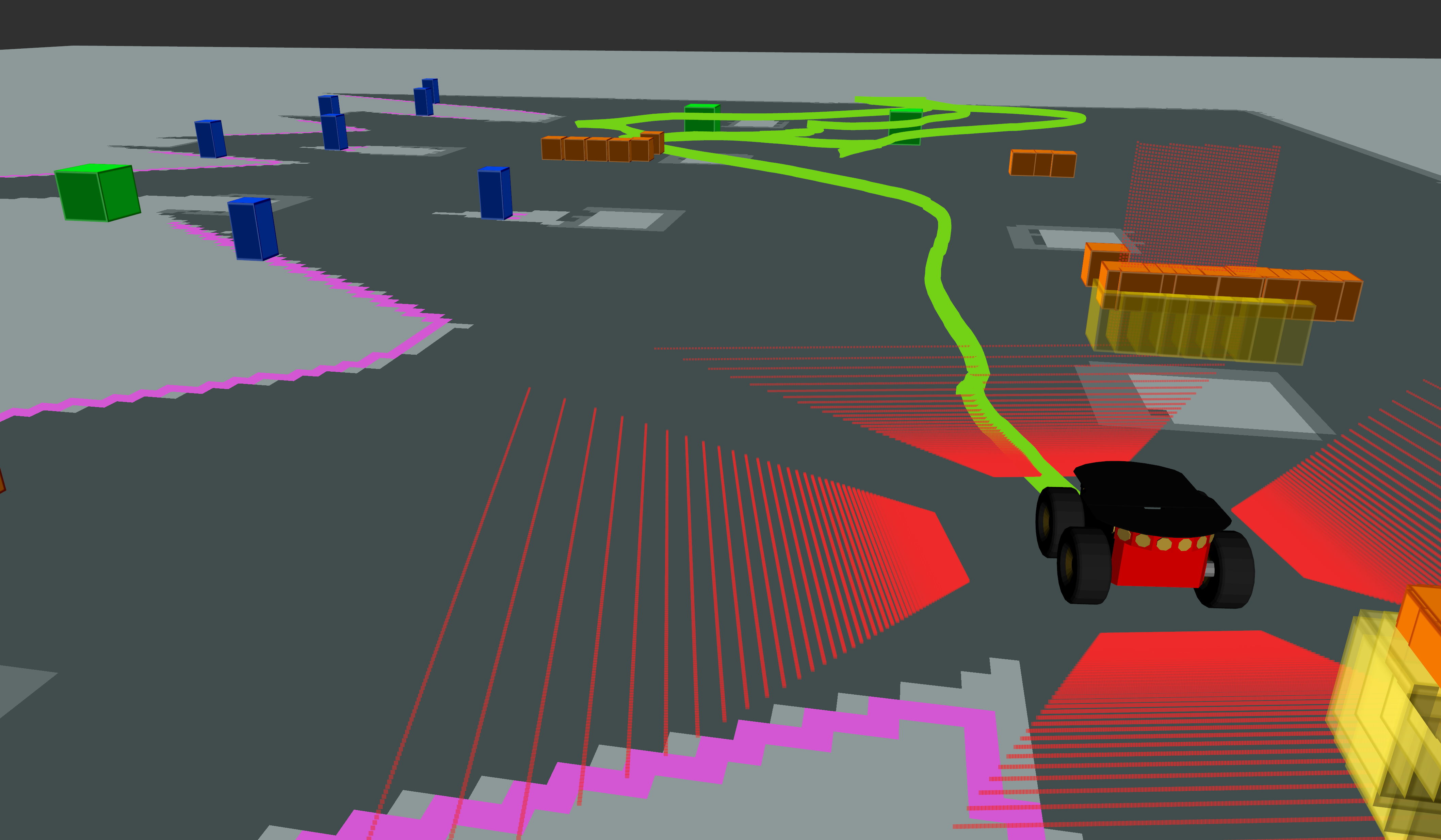}
    \end{subfigure}
    \par\bigskip
    \begin{subfigure}[b]{0.375\columnwidth}
        \includegraphics[width=\textwidth]{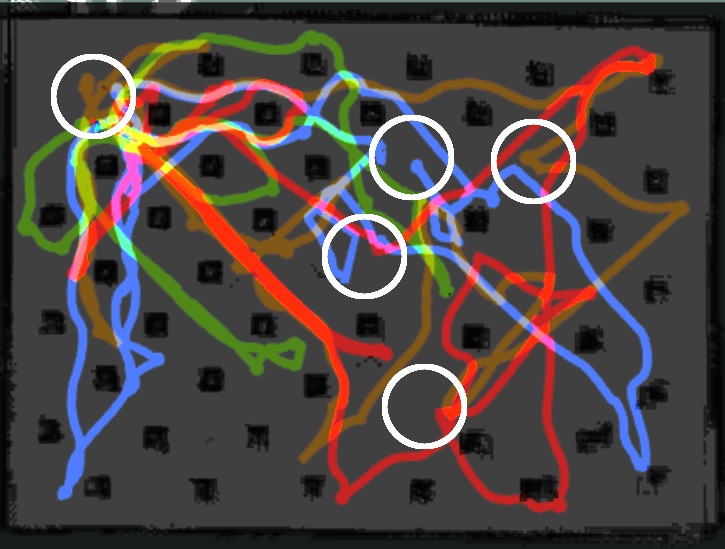}
    \end{subfigure}
    \hspace{0.5em}
    \begin{subfigure}[b]{0.375\columnwidth}
        \includegraphics[width=\textwidth]{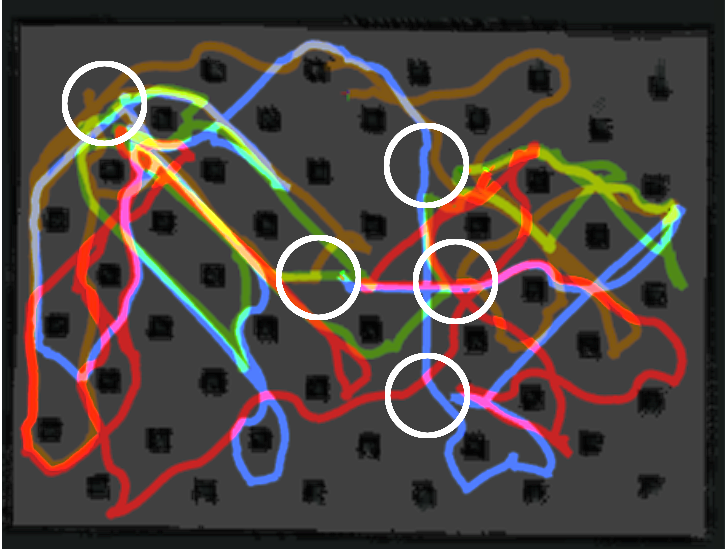}
    \end{subfigure}
    
    \caption{Rviz view of one robot (top) and the robots' trajectories (bottom). In the Rviz view, blue boxes represent frontiers and green boxes depict other robots. The trajectories plot below displays white circles indicating rendezvous locations, with colored lines representing the paths traversed by the robots.}
    \label{fig:trajectories}
\end{figure}

\section{Gazebo Simulation Results}

Our proof-of-concept addresses all aspects of the exploration mission using a mock communication model based on ROS network infrastructure as described by the robots' components in Fig.~\ref{fig:execution}.
We evaluated an exploration mission with 4 robots equipped with a 360-degree lidar for localization and mapping, and four RGB cameras short-range dynamic obstacle avoidance, each. The robots have a communication range of 3 meters.
Each robot calculates its paths using a roadmap-based method, employs an local planner to handle exploration and navigation uncertainties, and has an average traveling speed of 0.5 m/s.
For \textit{Simultaneous Localization and Mapping} (SLAM), each robot uses the gmapping package to solve SLAM individually and further shares their maps when appropriate.
Unlike the benchmark experiments, robots receive predetermined K and W matrices in the proof of concept, which is not detrimental to the purpose of evaluating the agreement execution subsystems integrated with a decentralized multi-robot exploration stack.
We evaluated the policy in two different trials and collected the robots' trajectories.
The exploration stack source code is available at: {\small\url{https://github.com/multirobotplayground/ROS-Noetic-Multi-robot-Sandbox}}.

Fig.~\ref{fig:trajectories} shows the execution of the method in two runs. We observed that our method can be deployed in a realistic multi-robot exploration stack with minor changes to existing controllers. With robust path and trajectory planners, the robots were able to fully explore the environment in the proposed testing area, executing agreements during the exploration process.
Furthermore, we observed the necessity of having a communication protocol to synchronize rendezvous events.
%
Despite this, our system was able to accomplish all posed challenges, highlighting the adaptability of our method in realistic scenarios. It successfully achieved the mission objectives of exploring the environment and maintaining intermittent connectivity under navigation and exploration uncertainties with communication constraints.



\section{Conclusion}
\label{sec:con}

This paper proposes a multi-robot frontier-based exploration method with intermittent rendezvous. 
We model the problem as a DEC-POMDP to maximize the utility of visiting frontiers while maintaining intermittent communication across the assigned mission area.
In our method, robots perform rendezvous through agreements at locations spread dynamically through the environment.
We provide the means to generate the rendezvous plan automatically by transforming the MRE problem into an instance of the JSSP and optimizing it, given its metrics.
Our results suggest that our method can explore faster with higher DEC-POMDP rewards without a base station or a relay network.
Further research involves explicitly incorporating uncertainties into the JSSP to reduce the time robots wait at rendezvous locations, incorporating a heterogeneous team of robots, and considering robots with energetic constraints and trajectories.


\bibliographystyle{IEEEtran}
\bibliography{IEEEabrv, bib/exploration, bib/hazards, bib/rescue, bib/communication, bib/opt, bib/connectivity, bib/schedule_jssp, bib/benchmarks, bib/evo, bib/decpomdp}

\end{document}